%% file: NormProp.tex
\documentclass{article}
\usepackage[titletoc,title]{appendix}
\usepackage{times}
\usepackage{graphicx} 
\usepackage{subfigure} 

\usepackage{natbib}

\usepackage{algorithm}
\usepackage{algorithmic}

\usepackage{hyperref}
\usepackage[fleqn]{amsmath}
\usepackage{amsthm}
\usepackage{amssymb}

\newtheorem{remark}{Remark}

\newtheorem{proposition}{Proposition}

\DeclareMathOperator{\tr}{tr}
\DeclareMathOperator{\diag}{diag}
\DeclareMathOperator{\var}{var}

\usepackage[accepted]{icml2016}

\icmltitlerunning{Normalization Propagation}

\begin{document} 

\twocolumn[
\icmltitle{Normalization Propagation: A Parametric Technique for Removing Internal Covariate Shift in Deep Networks }

\icmlauthor{Devansh Arpit}{devansha@buffalo.edu}
\icmlauthor{Yingbo Zhou}{yingbozh@buffalo.edu}
\icmlauthor{Bhargava U. Kota}{buralako@buffalo.edu}
\icmlauthor{Venu Govindaraju}{govind@buffalo.edu}
\icmladdress{SUNY Buffalo}

\icmlkeywords{Normalization Propagation, Internal Covariate Shift}

\vskip 0.3in
]

	
	%
	%

\begin{abstract} 
While the authors of Batch Normalization (BN) identify and address an important problem involved in training deep networks-- \textit{Internal Covariate Shift}-- the current solution has certain drawbacks. Specifically, BN depends on batch statistics for layerwise input normalization during training which makes the estimates of mean and standard deviation of input (distribution) to hidden layers inaccurate for validation due to shifting parameter values (especially during initial training epochs). Also, BN cannot be used with batch-size $ 1 $ during training. We address these drawbacks by proposing a non-adaptive normalization technique for removing internal covariate shift, that we call \textit{Normalization Propagation}. Our approach does not depend on batch statistics, but rather uses a data-independent parametric estimate of mean and standard-deviation in every layer thus being computationally faster compared with BN. We exploit the observation that the pre-activation before Rectified Linear Units follow Gaussian distribution in deep networks, and that once the first and second order statistics of any given dataset are normalized, we can forward propagate this normalization without the need for recalculating the approximate statistics for hidden layers. 
\end{abstract} 

\section{Introduction and Motivation}
\label{sec_intro}
\citet{bn} identified an important problem involved in training deep networks, \textit{viz.}, Internal Covariate Shift. It refers to the problem of shifting distribution of the input of every hidden layer in a deep neural network. This idea is borrowed from the concept of covariate shift \cite{covariate_shift}, where this problem is faced by a single input-output learning system. Consider the last layer of a deep network being used for classification; this layer essentially tries to learn $ P(Y | X) $, where $ Y $ is the class label random variable (r.v.) and $ X $ is the layer input r.v. However, learning a fixed $ P(Y | X) $ becomes a problem if $ P(X) $ changes continuously. As a result, this slows down training convergence.

Batch Normalization (BN) addresses this problem by normalizing the distribution of every hidden layer's input. In order to do so, it calculates the pre-activation mean and standard deviation using mini-batch statistics at each iteration of training and uses these estimates to normalize the input to the next layer. While this approach leads to a significant performance jump by addressing internal covariate shift, its estimates of mean and standard-deviation of hidden layer input for validation rely on mini-batch statistics, which are not representative of the entire data distribution (especially during initial training iterations). This is because the mini-batch statistics of input to hidden layers depends on the output from previous layers, which in turn depend on the previous layer parameters that keep shifting during training, and a moving average of these estimates are used for validation. Finally, due to involvement of batch statistics, BN is inapplicable with batch-size $ 1 $. 



In this paper, we propose a simple parametric normalization technique for addressing internal covariate shift that does not depend on batch statistics for normalizing the input to hidden layers and is less severely affected by the problem of shifting parameters during validation. In fact, we show that it is unnecessary to explicitly calculate mean and standard-deviation from mini-batches for normalizing the input to hidden layers even for training. Instead, a data independent estimate of these normalization components are available in closed form for every hidden layer, assuming the pre-activation values follow Gaussian distribution and that the weight matrix of hidden layers are roughly incoherent. We show how to forward propagate the normalization property (of the data distribution) to all hidden layers by exploiting the knowledge of the distribution of the pre-activation values (Gaussian) and some algebraic manipulations. Hence we call our approach \textit{Normalization Propagation}. 

\section{Background}
\label{sec_cov_shift}
It has long been known in Deep Learning community that input whitening and decorrelation helps in speeding up the training process. In fact, it is explicitly mentioned in \cite{lecun2012efficient} that this whitening should be performed before every layer so that the input to the next layer has zero mean. From the perspective of Internal Covariate Shift, what is required for the network to learn a hypothesis $ P(Y|X) $ at any given layer at every point in time during training, is for the distribution $ P(X) $ of the input to that layer to be fixed. While whitening could be used for achieving this task at every layer, it would be a very expensive choice (cubic order of input size) since whitening dictates computing the Singular Value Decomposition (SVD) of the input data matrix. However, \citet{desjardins2015natural} suggest to overcome this problem by approximating this SVD by: a) using sub-sampled training data to compute this SVD; b) computing it every few number of iteration and relying on the assumption that this SVD approximately holds for the iterations in between. In addition, each hidden layer's input is then \textit{whitened} by re-parametrizing a subset of network parameters that are involved in gradient descent. As mentioned in \citet{bn}, this re-parametrizing may lead to effectively cancelling/attenuating the effect of the gradient update step since these two operations are done independently. 

Batch Normalization \cite{bn} addresses both the above problems. First, they propose a strategy for normalizing the data at hidden layers such that the gradient update step accounts for this normalization. Secondly, this normalization is performed for units of each hidden layer independently (thus avoiding whitening) using mini-batch statistics. Specifically, this is achieved by normalizing the pre-activation $ \mathbf{u} = \mathbf{W}^{T}\mathbf{x} $ of all hidden layers as,
\begin{equation}
\hat{u}_{i} = \frac{u_{i} - \mathbb{E}_{\mathcal{B}}[u_{i}]}{\sqrt{\var_{\mathcal{B}}(u_{i})}}
\end{equation} 
where $ u_{i} $ denotes the $ i^{th} $ element of $ \mathbf{u} $ and the expectation/variance is calculated over the training mini-batch $ \mathcal{B} $. Notice since $ \mathbf{W} $ is a part of this normalization, it becomes a part of the gradient descent step as well. However, a problem common to both the above approaches is that of shifting network parameters upon which their approximation of input normalization for hidden layers depends.

\section{Normalization Propagation (\textit{NormProp}) Derivation}
We will now describe the idea behind NormProp. At a glance the problem at hand seems cyclic because estimating the mean and standard deviation of the input distribution to any hidden layer requires the input distribution of its previous layer (and hence its parameters) to be fixed to the optimal value before hand. However, as we will now show, we can side-step this naive approach and get an approximation of the true unbiased estimate using the knowledge that the pre-activation to every hidden layer follows a Gaussian distribution and some algebraic manipulation over the properties of the weight matrix. For the derivation below, we will focus on networks with ReLU activation, and later discuss how to extend our algorithm to other activation functions.
\subsection{Data Normalization}
Real world data generally follows Gaussian like distribution. Therefore, consider a data distribution $ \mathcal{X} $ in $ \mathbb{R}^{n} $ such that all the samples are normalized, \textit{i.e.}, 
\begin{equation}
\begin{split}
\mathbb{E}_{\mathbf{x}\in \mathcal{X}}[\mathbf{x}] = \mathbf{0}\\
\mathbb{E}_{\mathbf{x}\in \mathcal{X}}[x_{j}^{2}] =1 \mspace{5mu} \forall \mspace{5mu} j \in \{1,\hdots ,n\}
\end{split}
\end{equation}
Then our goal is to find a way to propagate this normalization to all the hidden layers without the need of explicit data dependent normalization. Depending on whether this input is passed through a convolutional layer or a fully connected layer, a part of the input or the entire input gets multiplied to a weight matrix. Irrespective of the case, lets use $ \mathbf{x} $ to denote this input for ease of notation; which can thus be the entire data vector or a subset of its element depending on the case. The pre-activation is given by $ \mathbf{u}\triangleq\mathbf{W}\mathbf{x} $, where $ \mathbf{W}\in \mathbb{R}^{m \times n} $ and $ m $ is the number of filters (we will ignore bias for now). As also claimed by \citet{bn,hyvarinen2000independent}, we assume the pre-activation ($ \mathbf{u} $) has a Gaussian form. So if we ensure that the pre-activation of the first hidden layer is a normalized Gaussian, then the hidden layer's output ($ \mbox{ReLU}( \mathbf{u} ) $) will be a Rectified Gaussian distribution. As mentioned in section \ref{sec_cov_shift}, we can choose to directly normalize the post-activation output $ \mbox{ReLU}( \mathbf{u} ) $. However, as we will now show, it is easier to find closed form estimates for normalizing $ \mathbf{u} $ instead.

\subsection{Mean and Standard-deviation Normalization for First Hidden Layer}
Clearly, since the input data $ \mathbf{x} $ is mean subtracted, the pre-activation to the first hidden layer $ \mathbf{u} $ also has zero mean from linearity,\textit{i.e.}, $ \mathbb{E}_{\mathbf{x}\in \mathcal{X}}[\mathbf{u}]=\mathbf{0} $. Now we want to ensure the variance of $ \mathbf{u} $ is $ 1 $. Let the covariance matrix of $ \mathbf{u} $ be denoted by $ \mathbf{\Sigma} $. Then the following proposition bounds how far $ \mathbf{\Sigma} $ is from a canonical distribution.
\begin{proposition}
	\label{prop_elliptical_data}
	(\textit{Canonical Error Bound}) Let $ \mathbf{u} = \mathbf{W}\mathbf{x} $ where $ \mathbf{x} \in \mathbb{R}^{n}$ and $ \mathbf{W} \in \mathbb{R}^{m \times n} $ such that $ \mathbb{E}_{\mathbf{x}}[\mathbf{x}]=\mathbf{0} $ and $ \mathbb{E}_{\mathbf{x}}[\mathbf{x}\mathbf{x}^{T}]=\sigma^{2}\mathbf{I} $ ($ \mathbf{I} $ is the identity matrix) . Then the covariance matrix of $ \mathbf{u} $ is approximately canonical satisfying, 
			\begin{align}
			\label{eq_sperical_data}
			\nonumber
			\min_{\mathbf{\alpha} } \lVert \mathbf{\Sigma} - \diag{(\mathbf{\alpha})}  \rVert_{F} \\
			\leq \sigma^{2} \mu\sqrt{ \sum_{i,j=1;i \neq j}^{m}   \lVert \mathbf{W}_{i} \rVert_{2}^{2} \lVert \mathbf{W}_{j} \rVert_{2}^{2} }
			\end{align}
		where $ \mathbf{\Sigma} = \mathbb{E}_{\mathbf{u}}[(\mathbf{u}-\mathbb{E}_{\mathbf{u}}[\mathbf{u}])(\mathbf{u}-\mathbb{E}_{\mathbf{u}}[\mathbf{u}])^{T}] $ is the covariance matrix of $ \mathbf{u} $, $ \mu $ is the coherence\footnote{\scriptsize Coherence is defined as $\max_{\mathbf{W}_{i}, \mathbf{W}_{j}, i\neq j} \frac{\lvert \mathbf{W}_{i}^{T}\mathbf{W}_{j} \rvert}{\lVert \mathbf{W}_{i} \rVert_{2} \lVert \mathbf{W}_{j} \rVert_{2}}$} of the rows of $ \mathbf{W} $, $ \mathbf{\alpha} \in \mathbb{R}^{m} $ is the closest approximation of the covariance matrix to a canonical ellipsoid and $ \diag(.) $ diagonalizes a vector to a diagonal matrix. The corresponding optimal $\alpha_{i}^{*} = \sigma^{2}\lVert \mathbf{W}_{i} \rVert_{2}^{2}$ $ \forall i \in \{ 1, \hdots, m\}  $.
\end{proposition}
The above proposition tells us two things. First, the covariance matrix of the pre-activation $ \mathbf{\Sigma} $ is approximately canonical (diagonal covariance matrix) if the above error bound can be made tight, and that this tightness can be controlled by certain properties of the weight matrix $ \mathbf{W} $. Second, if we want to normalize each element of the vector $ \mathbf{u}$ to have unit standard-deviation, then our best bet is to divide each $ \mathbf{u}_{i} $ by the corresponding weight length $ \lVert \mathbf{W}_{i} \rVert_{2}$ if we ensure tight canonical error bound. This is because the closest estimate of a diagonal variance for $ \mathbf{\Sigma} $ is $ \alpha_{i}^{*} = \lVert \mathbf{W}_{i} \rVert_{2}^{2}$ ($ \sigma=1 $ in our case). 

For any dictionary (weight matrix $ \mathbf{W} $), the bound above can be made tighter my minimizing coherence $ \mu $. In our approach, we also need to normalize each element of the vector $ \mathbf{u}$ to have unit standard-deviation which is achieved by dividing each $ u_{i} $ by $ \lVert \mathbf{W}_{i} \rVert_{2}$. Notice this automatically makes each hidden weight vector to effectively have unit $ \ell^{2} $ length. As a result, the error bound only depends on the coherence of $ \mathbf{W} $. On the other hand, it is generally observed that useful filters that constitute a good representation of real world data are roughly incoherent \cite{src,ksparse_ae}; thus ensuring the R.H.S is small thereby minimizing the error bound.

At this point, we have normalized the pre-activation $ \mathbf{u} $ to have zero mean and unit variance (divide each pre-activation element by corresponding $ \lVert \mathbf{W}_{i} \rVert_{2}$). As a result, the output of the first hidden layer ($ \mbox{ReLU}(\mathbf{u}) $) is Rectified Gaussian distribution. Notice that the above bound ensures the dimensions of $ \mathbf{u} $ and hence ($ \mbox{ReLU}(\mathbf{u}) $) are roughly uncorrelated. Thus, if we subtract the distribution mean from $ \mbox{ReLU}(\mathbf{u}) $ and divide by its standard deviation, we will have reduced the dynamics of the second layer to be identical to that of the first layer. The mean and standard deviation of the aforementioned Rectified Gaussian is,
\begin{remark}
	\label{prop_rec_gauss}
	\textit{(Post-ReLU distribution)} Let $ X \sim \mathcal{N}(0,1) $ and $ Y = \max(0,X) $. Then $ \mathbb{E}[Y] = \frac{1}{\sqrt{2\pi}} $ and $ \var(Y) = \frac{1}{{2}}\left(1-\frac{1}{\pi} \right) $
	\vspace{-9pt}
\end{remark}

Hence in order to normalize the post-activation $ \mbox{ReLU}(\mathbf{u}) $ to have zero mean and unit standard, the above calculated values can be used. Finally, in the case of Pooling (in Conv-Nets), we essentially take a block of post-activated units and take average or maximum of these values. If we consider each such unit to be independent then the distribution after pooling, will have a different mean and standard deviation. However, in reality, each of these units are highly correlated since they involve computation over either overlapping or spatially close patches. Therefore, we found that the distribution statistics do not get affected significantly and hence we do not recompute mean and standard deviation post-pooling.

\subsection{Propagation to Higher Layers}
With the above two operations, the dynamics of the second hidden layer become identical to that of the first hidden layer. By induction, repeating these two operations for every layer, \textit{viz.}--1) divide every hidden layer's pre-ReLU-activation by its corresponding $ \lVert \mathbf{W}_{i} \rVert_{2}$, where $ \mathbf{W} $ is the corresponding layer's weight matrix, 2) subtract and divide $ \sqrt{1/2\pi} $ and $ \sqrt{\frac{1}{{2}}\left(1-\frac{1}{\pi} \right)} $ (respectively) from every hidden layer's post-ReLU-activation-- we ensure that the input to every layer is a canonical distribution. While training, all these normalization operations are back-propagated. 

\subsection{Effect of NormProp on Jacobian of Hidden Layers}
\label{sec_jacobian_analysis}
It has been discussed in \citet{saxe2013exact,bn} that Jacobian of hidden layers with singular values close to one improves training convergence in deep networks. While BN has been shown to intuitively achieve this condition, we will now show more rigorously that NormProp (approximately) indeed achieves this condition. 

Let $ l \in \mathbb{R}^{m} $ be a vector such that the $ i^{th} $ element of $ l $ is given by $ l_{i} = 1/\lVert \mathbf{W}_{i}\rVert_{2} $. The output of a hidden unit using NormProp is given by $ \mathbf{o} = \frac{1}{c_{1}}\mbox{ReLU}( (\mathbf{W\mathbf{x}}) \odot l ) - c_{2}/c_{1} $ where $ \mathbf{W}\in \mathbb{R}^{m \times n} $, $ \mathbf{x} \in \mathbb{R}^{n} $, $ c_{1} = \sqrt{\frac{1}{{2}}\left(1-\frac{1}{\pi} \right)} $ and $ c_{2} = \frac{1}{\sqrt{2\pi}} $. Let $ \tilde{\mathbf{W}} $ be such that the $ i^{th} $ row of $ \tilde{\mathbf{W}} $ equals $ \mathbf{W}_{i}/\lVert \mathbf{W}_{i}\rVert_{2} $. Thus the output can be rewritten as $ \mathbf{o} = \frac{1}{c_{1}} \mbox{ReLU}(\tilde{\mathbf{W}}\mathbf{x}) -c_{2}/c_{1}$. Let $ \mathbf{J} $ denote the Jacobian of this output with respect to the previous layer input $ \mathbf{x} $. Then the $ i^{th} $ row of $ \mathbf{J} $ is given by
\begin{equation}
\begin{split}
\mathbf{J}_{i} \triangleq \frac{1}{c_{1}}\frac{\partial \mbox{ReLU}(\tilde{\mathbf{W}}_{i}\mathbf{x})}{\partial \tilde{\mathbf{W}}_{i}\mathbf{x}} \frac{\partial \tilde{\mathbf{W}}_{i}\mathbf{x}}{\partial \mathbf{x}} \\
= \frac{1}{c_{1}} \frac{\partial \mbox{ReLU}(\tilde{\mathbf{W}}_{i}\mathbf{x})}{\partial \tilde{\mathbf{W}}_{i}\mathbf{x}}\tilde{\mathbf{W}}_{i}
\end{split}
\end{equation}
where $ \tilde{\mathbf{W}}_{i} $ denotes the $ i^{th} $ row of $ \tilde{\mathbf{W}} $. Let $ \mathbf{1}_{\mathbf{x}} \in \mathbb{R}^{n}$ be an indicator vector such that the $ i^{th} $ element of $ \mathbf{1}_{\mathbf{x}} $ is given by
\begin{equation}
\begin{split}
 \mathbf{1}_{\mathbf{x}_{i}}  \triangleq \frac{\partial \mbox{ReLU}(\tilde{\mathbf{W}}_{i}\mathbf{x})}{\partial \tilde{\mathbf{W}}_{i}\mathbf{x}} = \mathbf{1}(\tilde{\mathbf{W}}_{i}\mathbf{x}>0)
\end{split}
\end{equation}
where $ \mathbf{1}(.) $ is the indicator operator. Let $ \mathbf{M}_{\mathbf{1}_{\mathbf{x}}} \in \mathbb{R}^{m \times n}$ be a matrix such that every column of $ \mathbf{M}_{\mathbf{1}_{\mathbf{x}}} $ is occupied by the vector $ \mathbf{1}_{\mathbf{x}}$. Then the entire Jacobian matrix can be written as $ \mathbf{J} =\frac{1}{c_{1}} (\mathbf{M}_{\mathbf{1}_{\mathbf{x}}} \odot \tilde{\mathbf{W}}) $. In order to analyze the singular values of $ \mathbf{J} $, we want to calculate $ \mathbf{J}\mathbf{J}^{T} $. From proposition \ref{prop_elliptical_data}, the covariance matrix $ \mathbf{\Sigma} $ of the pre-activation $ \tilde{\mathbf{W}}\mathbf{x} $ is given by $ \mathbf{\Sigma} = \sigma \tilde{\mathbf{W}}\tilde{\mathbf{W}}^{T} $, where $ \sigma=1$. Since the length of each $ \tilde{\mathbf{W}}_{i} $ is $ 1 $, $ \mathbf{\Sigma} $, and (therefore) $ \tilde{\mathbf{W}}\tilde{\mathbf{W}}^{T} $ is approximately an identity matrix if the rows of $ \tilde{\mathbf{W}} $ are incoherent. Thus,
\begin{equation}
\begin{split}
\mathbf{J}\mathbf{J}^{T} = \frac{1}{c_{1}^{2}}(\mathbf{M}_{\mathbf{1}_{\mathbf{x}}} \odot \tilde{\mathbf{W}})(\tilde{\mathbf{W}}^{T} \odot \mathbf{M}_{\mathbf{1}_{\mathbf{x}}}^{T})\\
\approx \frac{1}{c_{1}^{2}}\diag(\mathbf{1}_{\mathbf{x}} \odot \mathbf{1}_{\mathbf{x}}) = \frac{1}{c_{1}^{2}}\diag(\mathbf{1}_{\mathbf{x}})
\end{split}
\end{equation}
Finally taking an expectation of $ \mathbf{J}\mathbf{J}^{T} $ over the distribution of $ \mathbf{x} $ which is Normal, we get,
\begin{equation}
\begin{split}
\mathbb{E}_{\mathbf{x}}[\mathbf{J}\mathbf{J}^{T}] \approx  \frac{1}{c_{1}^{2}} \mathbb{E}_{\mathbf{x}}[\diag(\mathbf{1}_{\mathbf{x}})]\\
= \frac{1}{c_{1}^{2}}\int \diag(\mathbf{1}_{\mathbf{x}}) p(\mathbf{x})\mbox{d}(\mathbf{x})
\end{split}
\end{equation}
where $ p(.) $ denotes the density of $ \mathbf{x} $-- Normal distribution. From the definition of $ \mathbf{1}_{\mathbf{x}} $, it is straight forward to see that the result of the integral is a matrix with its diagonal equal to a vector of $ 0.5 $, hence, $ \mathbb{E}_{\mathbf{x}}[\mathbf{J}\mathbf{J}^{T}] \approx  \frac{0.5}{c_{1}^{2}}\mathbf{I}  \approx 1.47 \mathbf{I}$, where $ \mathbf{I} $ is the identity matrix. Thus the singular values of the Jacobian are $ \sqrt{(1.47)} = 1.21 $ which, being close to $ 1 $, approximately achieves dynamical isometry \cite{saxe2013exact} and should thus prevent the problem of exploding or diminishing gradients while training deep networks suggesting faster convergence. In the next section, we will use this value during the practical implementation of NormProp for improving the Jacobian to be approximately $ 1 $.

\section{NormProp: Implementation Details}
We have all the ingredients required to filter out the steps for Normalization Propagation for training any deep neural network with ReLU activation though out its hidden layers. Like BN, NormProp can be used alongside any optimization algorithm (eg. Stochastic Gradient Descent with/without momentum) for training deep networks.

\subsection{Normalize Data}
\label{sec_data_norm}
Since the core idea of NormProp is to propagate the data normalization through hidden layers, we offer two alternative choices either one of which can be used for normalizing the input to a NormProp network. As we will describe, both options are justified in their respective scenario.

1. \textit{Global Data Normalization:} In cases when the entire dataset -- approximately representing the true data distribution -- is available at hand, we compute the global mean and standard deviation for each feature element. Then the first step for NormProp is to subtract element-wise mean calculated over the entire dataset from each sample. Similarly divide each feature element by the element-wise standard-deviation. Ideally it is required by NormProp that all input dimensions be statistically uncorrelated, a property achieved by whitening for instance, but we suggest element-wise normalization as an approximation since it is computationally cheaper. Notice this precludes the dilemma of what range the input should be scaled to before passing through the network.

2. \textit{Batch Data Normalization:} In many real world scenario, streaming data is available and thus it is not possible to compute an unbiased estimate of global mean and standard deviation at any given point in time. In such cases, we propose to instead batch-normalize every mini-batch training data fed to the network. Again, we perform the normalization of each feature element independently for computational purposes. Notice this normalization is only performed at the data level, all hidden layers are still normalized by the NormProp strategy which is not affected by shifting model parameters as compared to BN. Moreover,  \textit{Batch Data Normalization} also serves as a regularization since each data sample gets a different representation each time depending on the mini-batch it comes with. Thus by using the \textit{Batch Data Normalization} strategy we actually benefit from the regularization aspect of BN but also overcome its drawbacks by computing the hidden layer mean and standard-deviation without depending on batch statistics. Notice this strategy is most effective when the incoming data is well shuffled.

\subsection{Initialize Network Parameters}
We use Normalized Initialization \cite{glorot2010understanding} for setting the initial values of all the weight matrices, both fully connected and convolutional. Bias vectors are initialized to zeros and scaling vectors (described in the next subsection) can either be initialized to $ 1 $ or as described in the next subsection.
\vspace{-8pt}
\subsection{Propagate Normalization}
\label{sec_prop_norm}
Similar to BN, we also make use of gradient-based-learnable scaling and bias parameters $ \mathbf{\gamma} $ and $ \mathbf{\beta} $ during implementation. We will now describe our normalization in detail for both fully connected and convolutional layers.
\subsubsection{Fully Connected Layers}
Consider any fully connected layer characterized by a weight matrix $ \mathbf{W} \in \mathbb{R}^{m \times n} $, bias $ \mathbf{\beta} \in \mathbb{R}^{m}$, scaling $ \mathbf{\gamma} \in \mathbb{R}^{m}$, input $ \mathbf{x} \in \mathbb{R}^{n} $ and activation ReLU. Here $ m $ denotes the number of filters and $ n $ denotes the input dimension. Then without NormProp, the $ i^{th} $ output unit $ {o}_{i} $ of the hidden layer would traditionally be:
\begin{equation}
{o}_{i} = \mbox{ReLU}(\mathbf{W}_{i}^{T}\mathbf{x} + {\beta}_{i})
\end{equation}
Now in the case of NormProp, the output $ {o}_{i} $ becomes,
\begin{equation}
{o}_{i} = \frac{1}{\sqrt{\frac{1}{{2}}\left(1-\frac{1}{\pi} \right)}} \left[\mbox{ReLU}\left(\frac{\gamma_{i}(\mathbf{W}_{i}^{T}\mathbf{x})}{\lVert \mathbf{W}_{i} \rVert_{2}} + {\beta}_{i} \right) - \sqrt{\frac{1}{2\pi}} \right]
\end{equation}
Here we initialize each $ \gamma_{i} $ to $1/1.21 $ in order to make the Jacobian close to one as suggested by our analysis in section \ref{sec_jacobian_analysis} for ReLU activation. Thus we call this number the \textit{Jacobian factor}. We found this initializing using Jacobian factor helps training with larger learning rates without diverging. However, one can also choose to treat the initialization value as a hyper-parameter.

\subsubsection{Convolutional Layers}
Consider any convolutional layer characterized by a filter matrix $ \mathbf{W} \in \mathbb{R}^{m \times d \times h \times w}$, bias $ \mathbf{\beta} \in \mathbb{R}^{m}$, scaling $ \mathbf{\gamma} \in \mathbb{R}^{m}$, input $ \mathbf{x} \in \mathbb{R}^{d \times L \times B} $ and activation ReLU along with any arbitrary choice of stride-size. Here, $ m $ denotes the number of filters, $ d $--depth, $ h $--height, $ w $-- width for input/filters and L,B-- height and width of image. Then without NormProp, the $ i^{th} $ output feature map $ \mathbf{o}_{i} $ of the hidden layer using the $ i^{th} $ filter $ \mathbf{W}_{i} \in \mathbb{R}^{d \times h \times w} $ would traditionally be:

\begin{equation}
\mathbf{o}_{i} = \mbox{ReLU}(\mathbf{W}_{i}\mathbf{*}\mathbf{x} + {\beta}_{i})
\end{equation}
where $ \mathbf{*} $ denotes the convolution operation. Now in the case of NormProp, the output feature map $ \mathbf{o}_{i} $ becomes,
\begin{equation}
\mathbf{o}_{i} = \frac{1}{\sqrt{\frac{1}{{2}}\left(1-\frac{1}{\pi} \right)}} \left[\mbox{ReLU}\left(\frac{\gamma_{i}(\mathbf{W}_{i}\mathbf{*}\mathbf{x})}{\lVert \mathbf{W}_{i} \rVert_{F}} + {\beta}_{i} \right) - \sqrt{\frac{1}{2\pi}} \right]
\end{equation}
where each element of $ {\gamma}_{i} $ is again initialized to $ 1/1.21 $. Notice each $ {\gamma}_{i} $  is multiplied to all outputs from the same corresponding filter and similarly all the scalars as well as the bias vector are broadcasted to all the dimensions. Pooling is done after this normalization process the same way as done traditionally.
\begin{figure}[t]
	\begin{tabular}{  c  }
		\includegraphics[width=0.9\columnwidth]{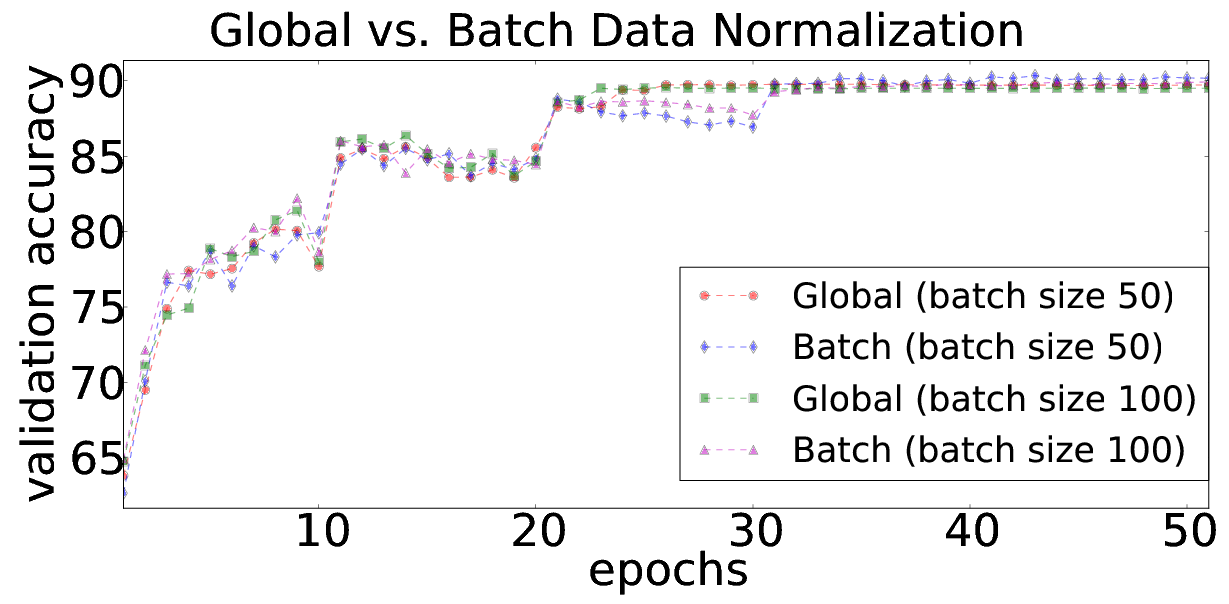}
	\end{tabular}
	\caption{Effect of Global vs. Batch Data Normalization on NormProp. Performance of NormProp is unaffected by the choice of data normalization strategy.\label{fig_global_vs_batch_data_norm}}

\end{figure}

\begin{figure*}
	\begin{minipage}[t]{0.33\textwidth}
		\includegraphics[width=1\columnwidth,trim=0.1in 0.in 0.1in 0.1in,clip]{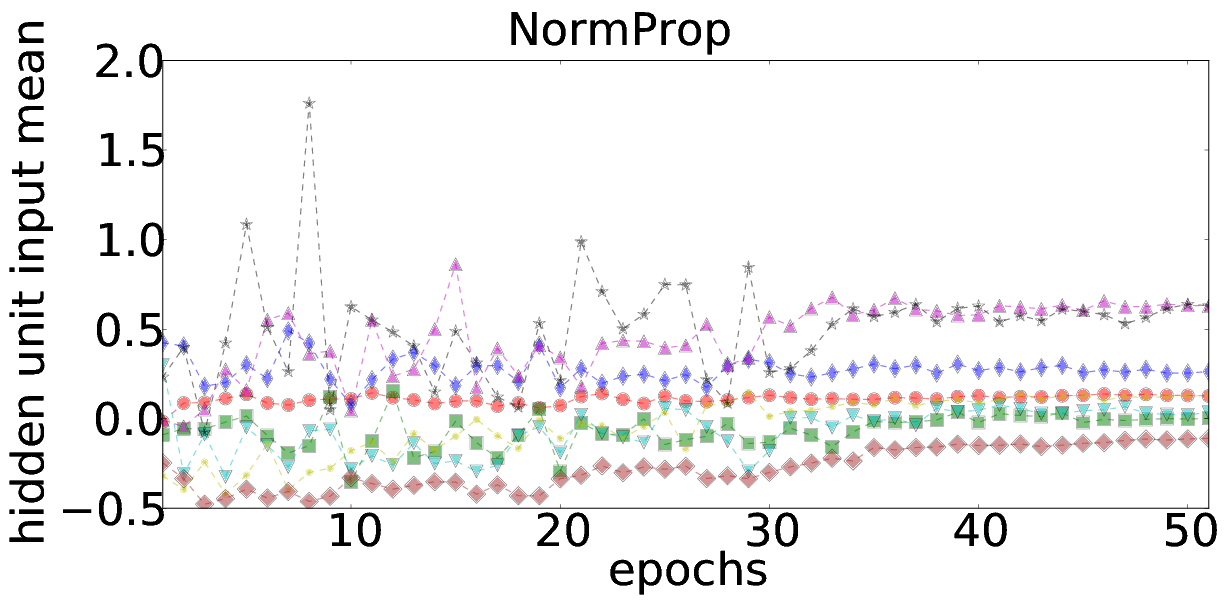}
	\end{minipage}
	\begin{minipage}[t]{0.33\textwidth}
		\includegraphics[width=1\columnwidth,trim=0.1in 0.in 0.1in 0.1in,clip]{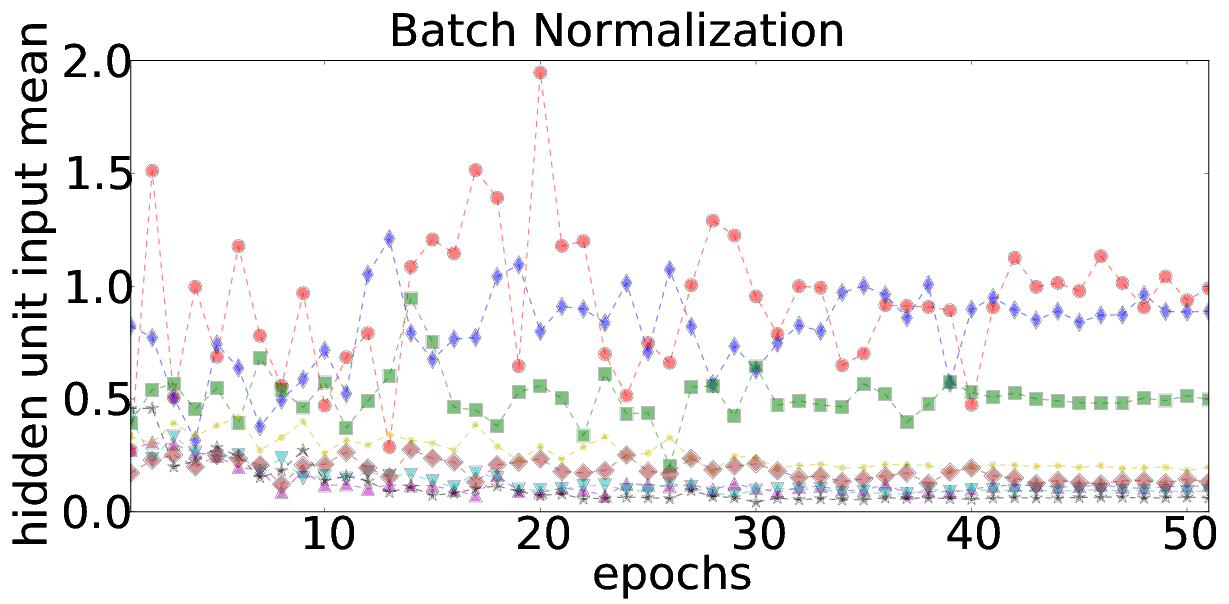}
		
	\end{minipage}
	\begin{minipage}[t]{0.33\textwidth}
		\includegraphics[width=1\columnwidth,trim=0.1in 0.in 0.1in 0.1in,clip]{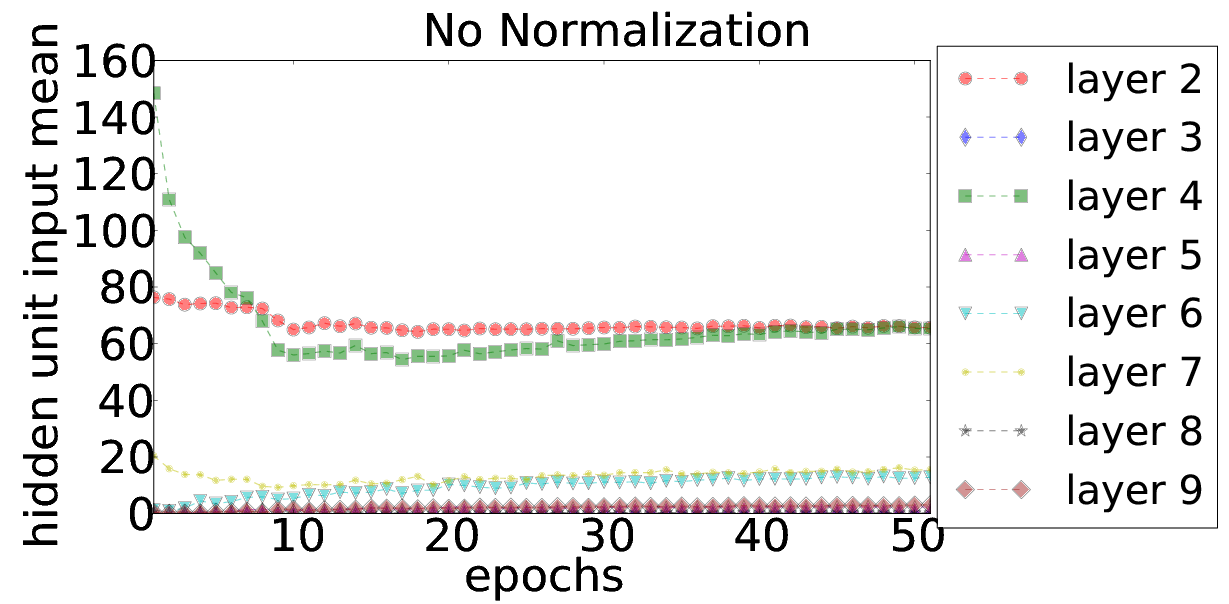}
		
	\end{minipage}
	\caption{Evolution of hidden layer input distribution mean (over validation set) of a randomly chosen unit for each hidden layer of a $ 9 $ layer convnet trained on CIFAR-$ 10 $. NormProp achieves both more stable distribution for lower layers as well as overall convergence closer to zero compared to BN. Hidden layer input distribution without normalization is incomparable due to extreme variations. \label{fig_hidden_input_mean}}
\end{figure*}

\subsection{Training}
\label{sec_training}
The network is trained using Back Propagation. While doing so, all the normalizations also get back-propagated at every layer. 

\textit{Optimization: }We use Stochastic Gradient Descent with momentum (set to $ 0.9 $) for training. Data shuffling also leads to performance improvement (this however, is true in general while training deep networks).

\textit{Learning Rate: }We found learning speeds up by reducing the learning rate by half whenever the training error starts saturating. Also, we found larger initial learning rate for larger batch size improves performance.

\textit{Constraints: }After every training iteration, we scale each hidden weight-vector/filter-map to have unit \textit{$ \ell^{2} $} length, \textit{i.e.}, we use \textit{$ \ell^{2} $} constraint on all hidden weights, both convolutional and fully connected. This is done because the scale of weight vectors do not affect network representation, so constraining the weights should reduce the parameter search space.

\textit{Regularizations: }We use weight decay along with the loss function; we found a small coefficient value of $ 0.0005-0.005 $ is necessary during training. We found \textit{Dropout} does not help during training; we believe this might be because Dropout changes the distribution of output of the layer it is applied, which affects NormProp.

\subsection{Validation and Testing}
Validation and test procedures are identical for NormProp. While validation/testing, each sample is first normalized using mean and standard deviation which are calculated depending on how the train data is normalized during training. In case we use \textit{Global Data Normalization} during training, we simply use the same global estimate of mean and standard deviation to normalize each test/validation sample. On the other hand, if \textit{Batch Data Normalization} is used during training, a running estimate of mean and standard deviation is maintained during training which is then used to normalize every test/validation sample. Finally, the input is forward propagated though the network with learned parameters using the same strategy described in section \ref{sec_prop_norm}.

\subsection{Extension to other Activation Functions}
\label{sec_other_act}
Even though our paper shows how to overcome the problem of Internal Covariate Shift specifically for networks with ReLU activation throughout, we have in essence proposed a general framework for propagating normalization done at data level to all hidden layers. All that is needed for extending NormProp to other activation functions is to compute the distribution mean ($ c_{2} $) and standard deviation ($ c_{1} $) of output after the activation function of choice, similar to what is shown in remark \ref{prop_rec_gauss}. Thus the general form of output for any given activation $ \sigma(.) $ becomes\footnote{\scriptsize Using the appropriate Jacobian Factor allows the use of larger learning rate; however, NormProp works without it as well.} (shown for convolution layer as an example),
\begin{equation}
	\mathbf{o}_{i} = \frac{1}{c_{1}} \left[\sigma \left(\frac{\gamma_{i}(\mathbf{W}_{i}\mathbf{*}\mathbf{x})}{ \lVert \mathbf{W}_{i} \rVert_{F}} + {\beta}_{i} \right) - c_{2} \right]
\end{equation}

This activation can be both parameter based or fixed. For instance, a parameter based activation is Parametric ReLU (PReLU, \citet{prelu}) (with parameter $ a $) given by,
\begin{equation}
\text{PReLU}_{a}(x) = \left\{ \begin{array}{ll}
x & if x>0\\
ax &  if x \leq 0 \\
\end{array} 
\right.
\end{equation}
Then the post PReLU distribution statistics is given by,
\begin{remark}
	\label{rem_prelu}
	Let $ X \sim \mathcal{N}(0,1) $ and $ Y = \text{PReLU}_{a}(X) $. Then $ \mathbb{E}[Y] = (1-a)\frac{1}{\sqrt{2\pi}} $ and $ \var(Y) = \frac{1}{{2}}\left( (1+a^{2})  -\frac{(1-a)^{2}}{\pi} \right) $
\end{remark}
Notice the distribution mean and standard deviation depends on the parameter $ a $ and thus will be involved in the normalization process. In case of non-parameter based activations (eg. Tanh, Sigmoid), one can either choose to analytically compute the statistics (like we did for ReLU) or compute these values empirically by simulation since the input distribution to the activation is a fixed Normal distribution. Thus NormProp is a general framework which can be extended to any activation function of choice.
\section{Empirical Results and Observations}
We want to verify the following:  a) performance comparison of NormProp when using \textit{Global Data Normalization} vs. \textit{Batch Data Normalization}; b) NormProp alleviates the problem of Internal Covariate Shift more accurately compared to BN; c) thus, convergence stability of NormProp is better than BN; d) effect of batch-size on the behaviour of NormProp, especially batch-size $ 1 $ (BN not applicable). Finally we report classification result on various datasets using NormProp and BN.

\textbf{Datasets}: We use the following datasets, \\
1) CIFAR-$10 $ \cite{cifar}-- It consists of $60,000$ $32 \times 32$ real world color images in 10 classes split into $50,000$ train and $10,000$ test images. We use $ 5000 $ images from train set for validation and remaining for training.\\
2) CIFAR-$ 100 $-- It has the same number of train and test samples as CIFAR-$ 10 $ but it has $ 100 $ classes. For training, we use hyperparameters same as those for CIFAR-$ 10 $.\\
3) SVHN \cite{svhn}-- It consists of $ 32 \times 32 $ color images of house numbers collected by Google Street View. It has $ 73,257 $ train images, $ 26,032$ test images and an additional $ 5,31,131$ train images. Similar to the protocol in \cite{Goodfellow13maxoutnetworks}, we select 400 samples per class from the train set and 200 samples
per class from the extra set as validation and use the remaining images of the train and extra sets for training.

\textbf{Experimental Protocols} (For experiments in sections \ref{sec_global_batch_norm} through \ref{sec_batch_size}): We use CIFAR-$ 10 $ with the following Network in Network \cite{nin} architecture\footnote{\scriptsize We use the following shorthand for a) conv layer: C(number of filters, filter size, stride size, padding); b) pooling: P(kernel size, stride, padding, pool mode)} $C(192,5,1,2)-C(160,1,1,0)-P(3,2,1,\text{max})-C(96,1,1,0)-C(192,5,1,2)-C(192,1,1,0)-P(3,2,1,\text{avg})-C(192,1,1,0)-C(192,5,1,0)-C(192,1,1,2)-C(10,1,1,0)-P(8,8,0,\text{avg})$. For any specified initial learning rate, we reduce it by half every $ 10 $ epochs. We use Stochastic Gradient Descent with momentum $ 0.9$. We use test set during validation for convergence analysis.

\subsection{Global vs. Batch Data Normalization}
\label{sec_global_batch_norm}
Since we offer two alternate ways to normalize data (section \ref{sec_data_norm}) fed to a NormProp network, we evaluate both strategies with different batch sizes to see the difference in performance. We use batch sizes\footnote{\scriptsize Notice this batch size has nothing to do with the data normalization strategies in discussion. Different batch sizes are used only for adding more variation in experiments.} $ 50 $ and $ 100 $ using initial learning rates $ 0.05 $ and $ 0.08 $ respectively. The results are shown in figure \ref{fig_global_vs_batch_data_norm}. The performance \footnote{\scriptsize Even though the numbers are very close, the best accuracy of $ 90.35\% $ is achieved by Batch Data Normalization using batch size $ 50 $.} using both strategies is very similar for both batch sizes, converging in only $ 30 $ epochs. This shows the robustness and applicability of NormProp in both streaming data as well as block data scenario. However, since Batch Data Normalization strategy is a more practical choice, we stick to this strategy for the rest of the experiments when using NormProp.

\subsection{NormProp \textit{vs.} BN-- Internal Covariate Shift}
\label{sec_int_cov_shift}
The fundamental goal of our paper (as well as that of Batch Normalization \citealp{bn}) is to alleviate the problem of \textit{Internal Covariate Shift}. This implies preventing the distribution of hidden layer inputs from shifting while the network is being trained. In deep networks, the features generated by higher layers are completely dependent on the lower features since all the information in data is propagated from lower to higher layers. Thus the problem of Internal Covariate Shift in lower hidden layers is expected to affect the overall performance more severely as compared to the same problem in higher layers.

\begin{figure}
	\begin{minipage}{0.5\textwidth}
		\begin{tabular}{  c     }
			\includegraphics[width=0.83\columnwidth,trim=0.1in 0.1in 0.1in 0.1in,clip]{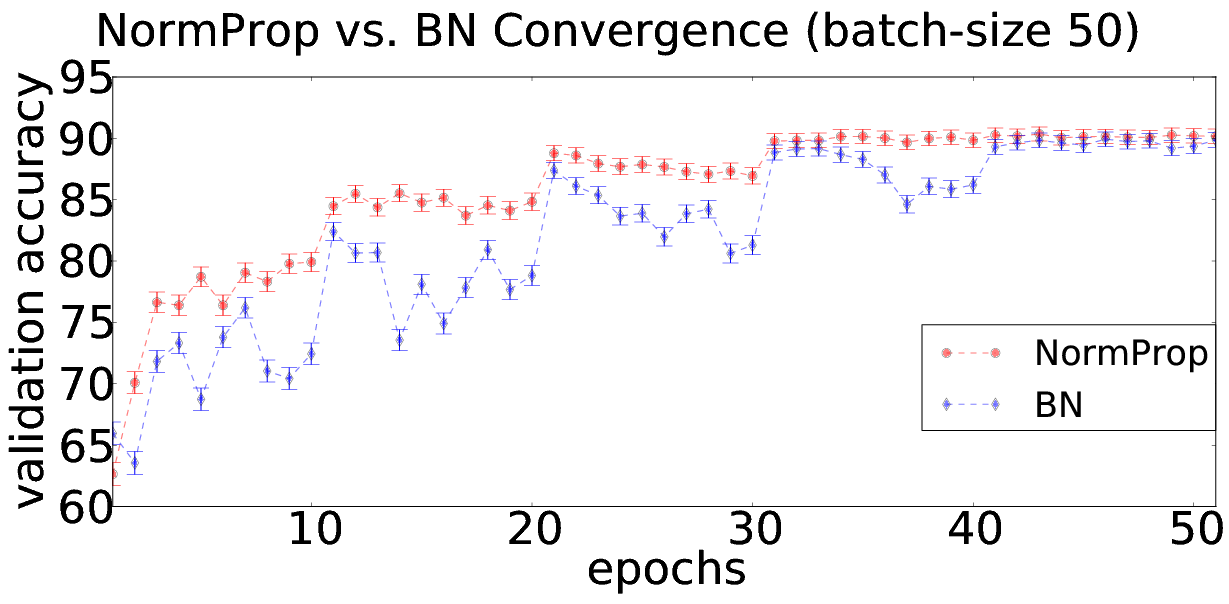}
		\end{tabular}
	\end{minipage}
	
	\caption{NormProp vs. BN convergence stability. Clearly NormpProp avoids Internal Covariate shift more accurately resulting in a more stable convergence (especially during initial training). Caps on markers show $ 95\% $ confidence interval.\label{fig_bn_np_convergence}}
\end{figure}

In order to study the effect of normalization by NormProp and BN on hidden layers, we train two separate networks using each strategy and an additional network without any normalization as our baseline. After every training epoch, we record the mean of the input distribution (over the validation set) to a single randomly chosen (but fixed) unit in each hidden layer. We use batch size $ 50 $ and an initial learning rate of $ 0.05 $ for NormProp and BN, and $ 0.0001 $ for training the network without any normalization (larger learning rates cause divergence). For the $ 9 $ layer convolutional networks we train, the input mean to the last $ 8 $ layers against training epoch are shown in figure \ref{fig_hidden_input_mean}. There are three important observations in these figures: a) NormProp achieves significantly more stable input distribution for lower hidden layers compared to BN, thus facilitating good lower level representation; b) the input distribution for all hidden layers converge after $ ~32 $ epochs for NormProp. On the other hand, the input distribution to the second layer for BN remains un-converged even after $ 50 $ epochs; c) on an average, input distribution to all layers converge closer to zero for NormProp (avg. $ 0.19 $) as compared to BN (avg. $ 0.33 $). Finally the performance of the network trained without any normalization is in-comparable to the normalized ones due to large variations in the hidden layer input distribution (especially the lower layers). This experiment also serves to show the \textit{Canonical Error Bound} (proposition \ref{prop_elliptical_data}) is small since the input statistics to hidden layers are roughly preserved.

\subsection{Convergence Stability of NormProp vs. BN}
\label{sec_convergence}
As a result of alleviating Internal Covariate Shift more accurately during validation as compared to BN, NormProp is expected to achieve a more stable convergence. We confirm this intuition by recording the validation accuracy while the network is being trained.  We use batch size $ 50 $ and initial learning rates $ 0.05 $. The plot is shown in figure \ref{fig_bn_np_convergence}\footnote{\scriptsize We observed identical trends on SVHN and CIFAR-$ 100 $. Additionally, we also experimented optimizing with SGD without momentum and RMS prop \cite{tieleman2012lecture}. We found in general (for most mini-batch sizes) the performance of RMS prop was worse than SGD with momentum while that of SGD without momentum was worse than both. On the other hand, RMSProp generally performed better than SGD but SGD with batch size 1 was very similar to SGD-Momentum.}. Clearly NormProp achieves a more stable convergence in general, but especially during initial training. This is because NormProp achieves a more stable  hidden layer input distribution computed for validation.

\begin{figure}
	\begin{tabular}{  c  }
		\includegraphics[width=0.83\columnwidth]{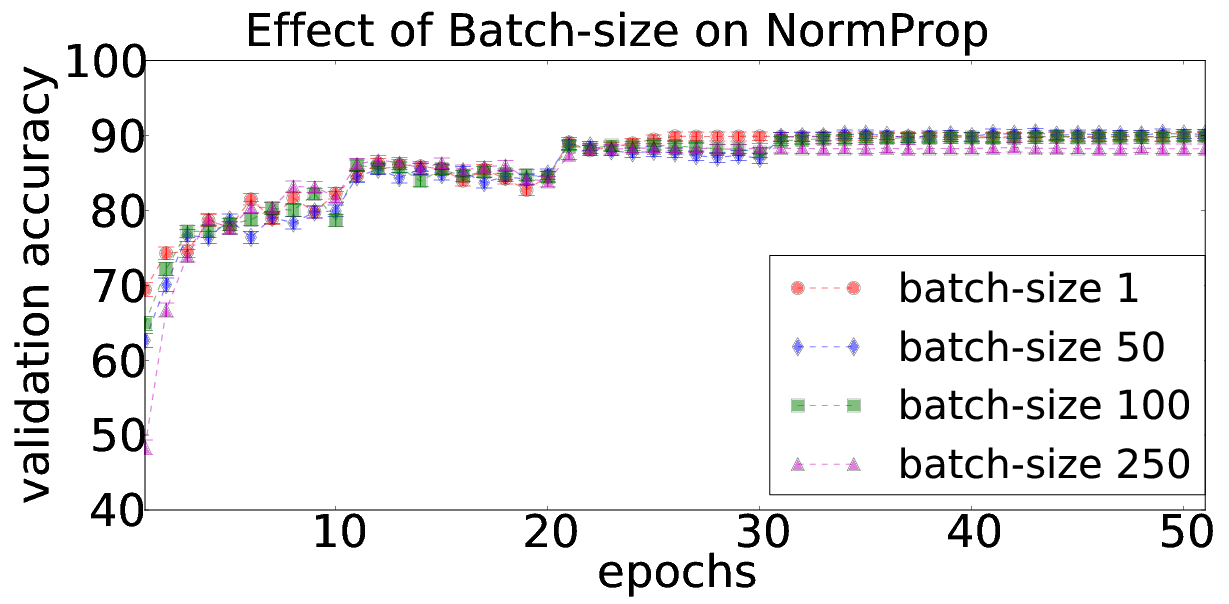}
	\end{tabular}
	\caption{Effect training batch-size on NormProp. NormProp achieves slightly better performance with decreasing batch-size. Caps on markers show $ 95\% $ confidence interval.\label{fig_effect_batch_size_normprop}}
\end{figure}

\subsection{Effect of Batch-size on NormProp}
\label{sec_batch_size}
We want to see the effect of batch-size used during training with NormProp. Since it is also possible to train with batch size $ 1 $ (using Global Data Normalization at data layer), we compare the validation performance of NormProp during training for various batch sizes including $ 1 $. The plots are shown in figure \ref{fig_effect_batch_size_normprop}. The performance of NormProp is largely unaffected by batch size although lower batch sizes seem to yield better performance.

\subsection{Results on various Datasets}
We evaluate NormProp and BN on CIFAR-$ 10 $, CIFAR-$ 100 $ and SVHN datasets, but also report existing state-of-the-art (SOTA) results. For all the datasets and both methods, we use the same architecture as mentioned in the experimental protocol above except for CIFAR-$ 100 $, the last convolutional layer is $ C(100,1,1,0) $ instead of $ C(10,1,1,0) $. For CIFAR datasets we use batch size $ 50 $ and an initial learning rate of $ 0.05 $ and reduce it by half after every $ 25 $ epochs and train for $ 200 $ epochs. Since SVHN is a much larger dataset, we only train for $ 25 $ epochs with batch size $ 100 $ and an initial learning rate of $ 0.08 $ and reduce it by half after every $ 5 $ epochs. We use Stochastic gradient descent with momentum ($ 0.9 $). For CIFAR-$ 10 $ and CIFAR-$ 100 $, we train using both without data augmentation and with data augmentation (horizontal flipping only); and no data augmentation for SVHN. We did not pre-process any of the datasets. The results are shown in table \ref{tab_classification}. We find NormProp consistently achieves either better or competitive performance compared to BN, but also beats existing SOTA results.
\begin{table}[t]
	\scriptsize
		\caption{Performance comparison of NormProp (ours) with Batch Normalization along with other State-of-the-art methods on various datasets.}
		\centering%
		\begin{tabular}{l c}
			Datasets and Methods & Test Error ($\%$) \tabularnewline
			\hline
			CIFAR-$ 10 $ & \tabularnewline
			\hline
			\multicolumn{2}{c}{without data augmentation} \tabularnewline
			\hline
			$ \mspace{20mu} $NormProp  & \textbf{9.11} \tabularnewline
			$ \mspace{20mu} $Batch Normalization   & 9.41 \tabularnewline
			$ \mspace{20mu} $NIN + ALP units \cite{agostinelli2014learning} & 9.59 \tabularnewline
			$ \mspace{20mu} $NIN \cite{nin} & 10.47 \tabularnewline
			$ \mspace{20mu} $DSN \cite{dsn} & 9.69 \tabularnewline
			$ \mspace{20mu} $Maxout \cite{Goodfellow13maxoutnetworks} & 11.68 \tabularnewline
			\hline
			\multicolumn{2}{c}{with data augmentation} \tabularnewline
			\hline
			$ \mspace{20mu} $NormProp  & 7.47 \tabularnewline
			$ \mspace{20mu} $Batch Normalization & \textbf{7.25} \tabularnewline
			$ \mspace{20mu} $NIN + ALP units \cite{agostinelli2014learning} & 7.51 \tabularnewline
			$ \mspace{20mu} $NIN \cite{nin} & 8.81 \tabularnewline
			$ \mspace{20mu} $DSN \cite{dsn} & 7.97 \tabularnewline
			$ \mspace{20mu} $Maxout \cite{Goodfellow13maxoutnetworks} & 9.38 \tabularnewline
			\hline
			CIFAR-$ 100 $  &  \tabularnewline
			\hline
			\multicolumn{2}{c}{without data augmentation} \tabularnewline
			\hline
			$ \mspace{20mu} $NormProp & \textbf{32.19} \tabularnewline
			$ \mspace{20mu} $Batch Normalization  & 35.32 \tabularnewline
			$ \mspace{20mu} $NIN + ALP units \cite{agostinelli2014learning} & 34.40 \tabularnewline
			$ \mspace{20mu} $NIN \cite{nin} & 35.68 \tabularnewline
			$ \mspace{20mu} $DSN \cite{dsn} & 34.57 \tabularnewline
			$ \mspace{20mu} $Maxout \cite{Goodfellow13maxoutnetworks} & 38.57 \tabularnewline
			\hline
			\multicolumn{2}{c}{with data augmentation} \tabularnewline
			\hline
			$ \mspace{20mu} $NormProp  & \textbf{29.24} \tabularnewline
			$ \mspace{20mu} $Batch Normalization  & 30.26 \tabularnewline
			$ \mspace{20mu} $NIN + ALP units \cite{agostinelli2014learning} & 30.83 \tabularnewline
			$ \mspace{20mu} $NIN \cite{nin} & - \tabularnewline
			$ \mspace{20mu} $DSN \cite{dsn} & - \tabularnewline
			$ \mspace{20mu} $Maxout \cite{Goodfellow13maxoutnetworks} & - \tabularnewline
			\hline
			SVHN & \tabularnewline
			\hline
			$ \mspace{20mu} $NormProp & \textbf{1.88} \tabularnewline
			$ \mspace{20mu} $Batch Normalization & {2.25} \tabularnewline
			$ \mspace{20mu} $NIN + ALP units \cite{agostinelli2014learning} & - \tabularnewline
			$ \mspace{20mu} $NIN \cite{nin} & 2.35 \tabularnewline
			$ \mspace{20mu} $DSN \cite{dsn} & 1.92 \tabularnewline
			$ \mspace{20mu} $Maxout \cite{Goodfellow13maxoutnetworks} & 2.47 \tabularnewline
			\hline
		\end{tabular}\label{tab_classification}
\vspace{-17pt}
\end{table}

\subsection{Training Speed}
Since there is no need for estimating the running average values of input mean and standard deviation for hidden layers for NormProp algorithm, it expected to be faster compared to Batch Normalization. So we record the time taken for NormProp and BN for 1 epoch of training on CIFAR-$ 10 $ dataset using the experimental protocol used for above experiments. On an NVIDIA GeForce GTX Titan X GPU with Intel i7-3930K CPU and 32GB Ram machine, NormProp takes $\sim 84$ sec while BN takes $ \sim 96 $ sec.

\section{Conclusion}
We have proposed a novel algorithm for addressing the problem of Internal Covariate Shift involved during training deep neural networks that overcomes certain drawbacks of Batch Normalization (BN). Specifically, we propose a parametric approach (NormProp) that avoids estimating the mean and standard deviation of hidden layers' input distribution using input data mini-batch statistics (that involve shifting network parameters). Instead, NormProp relies on normalizing the statistics of the given dataset and conditioning the weight matrix which ensures normalization done for the dataset is propagated to all hidden layers. Thus NormProp does not need to maintain a moving average estimate of batch statistics of hidden layer inputs for validation/test phase, thus being more representative of the entire data distribution (especially during initial training period when parameters change drastically). This also enables the use of batch size $ 1 $ for training. Although we have shown how to apply NormProp in detail for networks with ReLU activation, we have discussed (section \ref{sec_other_act}) how to extend it for other activations as well. We have empirically shown NormProp achieves more stable convergence and hidden layer input distribution over validation set during training, and better/competitive classification performance compared with BN while being faster by omitting the need to compute mini-batch estimate of mean/standard-deviation for hidden layers' input. In conclusion, our approach is applicable alongside any activation function and cost objectives for improving training convergence.

\bibliography{NormProp}
\bibliographystyle{icml2016}

\newpage
\input{appendix}
\end{document}

%% file: appendix.tex
\onecolumn
\appendix    
\appendixpage
\begin{appendix}
\setcounter{theorem}{0}
\setcounter{proposition}{0}
\setcounter{corollary}{0}
\setcounter{remark}{0}
\section{Proofs}
\begin{proposition}
	\label{prop_elliptical_data}
	Let $ \mathbf{u} = \mathbf{W}\mathbf{x} $ where $ \mathbf{x} \in \mathbb{R}^{n}$ and $ \mathbf{W} \in \mathbb{R}^{m \times n} $ such that $ \mathbb{E}_{\mathbf{x}}[\mathbf{x}]=\mathbf{0} $ and $ \mathbb{E}_{\mathbf{x}}[\mathbf{x}\mathbf{x}^{T}]=\sigma^{2}\mathbf{I} $ ($ \mathbf{I} $ is the identity matrix) . Then the covariance matrix of $ \mathbf{u} $ is approximately canonical satisfying, 
	\begin{equation}
	\label{eq_sperical_data}
	\min_{\mathbf{\alpha} } \lVert \mathbf{\Sigma} - \diag{(\mathbf{\alpha})}  \rVert_{F} \leq \sigma^{2}\mu\sqrt{  \sum_{i,j=1;i \neq j}^{m}   \lVert \mathbf{W}_{i} \rVert_{2}^{2} \lVert \mathbf{W}_{j} \rVert_{2}^{2} }
	\end{equation}
	where $ \mathbf{\Sigma} = \mathbb{E}_{\mathbf{u}}[(\mathbf{u}-\mathbb{E}_{\mathbf{u}}[\mathbf{u}])(\mathbf{u}-\mathbb{E}_{\mathbf{u}}[\mathbf{u}])^{T}] $ is the covariance matrix of $ \mathbf{u} $, $ \mu $ is the coherence of the rows of $ \mathbf{W} $, $ \mathbf{\alpha} \in \mathbb{R}^{m} $ is the closest approximation of the covariance matrix to a canonical ellipsoid and $ \diag(.) $ diagonalizes a vector to a diagonal matrix. The corresponding optimal $\alpha_{i}^{*} = \sigma^{2}\lVert \mathbf{W}_{i} \rVert_{2}^{2}$ $ \forall i \in \{ 1, \hdots, m\}  $.

%
	\begin{proof}
		Notice that,
		\begin{equation}
		\begin{split}
		\mathbb{E}_{\mathbf{u}}[\mathbf{u}] = \mathbf{W}\mathbb{E}_{\mathbf{x}}[\mathbf{x}] = \mathbf{0}
		\end{split}
		\end{equation}
		On the other hand, the covariance of $ \mathbf{u} $ is given by,
		\begin{equation}
		\begin{split}
		\mathbf{\Sigma} = \mathbb{E}_{\mathbf{u}}[(\mathbf{u}-\mathbb{E}_{\mathbf{u}}[\mathbf{u}])(\mathbf{u}-\mathbb{E}_{\mathbf{u}}[\mathbf{u}])^{T}] = \mathbb{E}_{\mathbf{x}}[(\mathbf{W}\mathbf{x}-\mathbf{W}\mathbb{E}_{\mathbf{x}}[\mathbf{x}])(\mathbf{W}\mathbf{x}-\mathbf{W}\mathbb{E}_{\mathbf{x}}[\mathbf{x}])^{T}]\\
		=\mathbb{E}_{\mathbf{x}}[\mathbf{W}(\mathbf{x}-\mathbb{E}_{\mathbf{x}}[\mathbf{x}])(\mathbf{x}-\mathbb{E}_{\mathbf{x}}[\mathbf{x}])^{T}\mathbf{W}^{T}]\\
		= \mathbf{W}\mathbb{E}_{\mathbf{x}}[(\mathbf{x}-\mathbb{E}_{\mathbf{x}}[\mathbf{x}])(\mathbf{x}-\mathbb{E}_{\mathbf{x}}[\mathbf{x}])^{T}]\mathbf{W}^{T}
		\end{split}
		\end{equation}
		Since $ \mathbf{x} $ has spherical covariance, the off-diagonal elements of $ \mathbb{E}_{\mathbf{x}}[(\mathbf{x}-\mathbb{E}_{\mathbf{x}}[\mathbf{x}])(\mathbf{x}-\mathbb{E}_{\mathbf{x}}[\mathbf{x}])^{T}] $ are zero and the diagonal elements are the variance of any individual unit, since all units are identical. Thus, 
		\begin{equation}
		\begin{split}
		\mathbb{E}_{\mathbf{u}}[(\mathbf{u}-\mathbb{E}_{\mathbf{u}}[\mathbf{u}])(\mathbf{u}-\mathbb{E}_{\mathbf{u}}[\mathbf{u}])^{T}] =\sigma^{2} \mathbf{W}\mathbf{W}^{T}
		\end{split}
		\end{equation}
		Thus,
		\begin{equation}
		\label{eq_shperical_bound_derivation}
		\begin{split}
		\lVert \mathbf{\Sigma} - \diag{(\mathbf{\alpha})}  \rVert_{F}^{2} = \tr \left( (\sigma^{2}\mathbf{W}\mathbf{W}^{T} - \diag{(\mathbf{\alpha})} )(\sigma^{2}\mathbf{W}\mathbf{W}^{T} - \diag{(\mathbf{\alpha})} )^{T} \right)\\
		= \tr \left( \sigma^{4}\mathbf{W}\mathbf{W}^{T}\mathbf{W}\mathbf{W}^{T} + \diag{(\mathbf{\alpha}^{2})}  - \sigma^{2}\diag{(\mathbf{\alpha})}  \mathbf{W}\mathbf{W}^{T} - \sigma^{2}  \mathbf{W}\mathbf{W}^{T}\diag{(\mathbf{\alpha})} \right)\\
		=  \sigma^{4}\lVert \mathbf{W}\mathbf{W}^{T} \rVert_{F}^{2} + \sum_{i=1}^{m} \left( \alpha_{i}^{2} - 2\sigma^{2}\alpha_{i} \lVert \mathbf{W}_{i} \rVert_{2}^{2} \right)  \\
		\leq  \sigma^{4} \sum_{i=1}^{m} \left( \lVert \mathbf{W}_{i} \rVert_{2}^{4} \right) + \sum_{i,j=1;i \neq j}^{m} \mu^{2} \lVert \mathbf{W}_{i} \rVert_{2}^{2} \lVert \mathbf{W}_{j} \rVert_{2}^{2} + \sum_{i=1}^{m} \left( \alpha_{i}^{2} - 2\sigma^{2}\alpha_{i} \lVert \mathbf{W}_{i} \rVert_{2}^{2} \right)
		\end{split}
		\end{equation}
		$ \mathbf{\alpha}^{2} $ in the above equation denotes element-wise square of elements of $ \alpha $.
		Finally minimizing w.r.t $ \alpha_{i}$ $ \forall i \in \{ 1, \hdots,m \} $, leads to $\alpha_{i}^{*} = \sigma^{2}\lVert \mathbf{W}_{i} \rVert_{2}^{2}$. Substituting this into equation \ref{eq_shperical_bound_derivation}, we get,
		\begin{equation}
		\begin{split}
		\lVert \mathbf{\Sigma} - \diag{(\mathbf{\alpha})} \rVert_{F}^{2} \leq \sigma^{4}  \sum_{i,j=1;i \neq j}^{m} \mu^{2} \lVert \mathbf{W}_{i} \rVert_{2}^{2} \lVert \mathbf{W}_{j} \rVert_{2}^{2} 
		\end{split}
		\end{equation}
		
	\end{proof}
\end{proposition}

\begin{remark}
	\label{prop_rec_gauss}
	Let $ X \sim \mathcal{N}(0,1) $ and $ Y = \max(0,X) $. Then $ \mathbb{E}[Y] = \frac{1}{\sqrt{2\pi}} $ and $ \var(Y) = \frac{1}{{2}}\left(1-\frac{1}{\pi} \right) $
	
\begin{proof}
	For the definition of $ X $ and $ Y $, we have,
	\begin{equation}
	\mathbb{E}[Y] = \frac{1}{2}.0 + \frac{1}{2}\mathbb{E}[Z] = \frac{1}{2}\mathbb{E}[Z]
	\end{equation}
	where $ Z $ is sampled from a Half-Normal distribution such that $ Z = |X| $; thus $ \mathbb{E}[Z] = \sqrt{\frac{2}{\pi}} $ leading to the claimed result. In order to compute variance, notice that $ \mathbb{E}[Y^{2}] = 0.5\mathbb{E}[Z^{2}] $. Then,
	\begin{equation}
	\begin{split}
	\var(Y) = \mathbb{E}[Y^{2}] - \mathbb{E}[Y]^{2} = 0.5\mathbb{E}[Z^{2}] - \frac{1}{4}\mathbb{E}[Z]^{2} = 0.5(\var(Z) + \mathbb{E}[Z]^{2})  - \frac{1}{4}\mathbb{E}[Z]^{2} 
	\end{split}
	\end{equation}
	Substituting $ \var(Z) =  1-\frac{2}{\pi}$ yields the claimed result.
\end{proof}
\end{remark}


%

\begin{remark}
	\label{rem_prelu}
	Let $ X \sim \mathcal{N}(0,1) $ and $ Y = \text{PReLU}_{a}(X) $. Then $ \mathbb{E}[Y] = (1-a)\frac{1}{\sqrt{2\pi}} $ and $ \var(Y) = \frac{1}{{2}}\left( (1+a^{2})  -\frac{(1-a)^{2}}{\pi} \right) $
	
	\begin{proof}
		For the definition of $ X $ and $ Y $, half the mass of $Y $ is concentrated on $ \mathbb{R}^{+} $ with Half-Normal distribution, while the other half of the mass is concentrated on $ \mathbb{R}^{-\text{sign}(a)} $with Half-Normal distribution scaled with $ |a| $. Thus,
		\begin{equation}
		\mathbb{E}[Y] = -\frac{a}{2}\mathbb{E}[Z] + \frac{1}{2}\mathbb{E}[Z] = (1 - a)\frac{1}{2}\mathbb{E}[Z]
		\end{equation}
		where $ Z $ is sampled from a Half-Normal distribution such that $ Z = |X| $; thus $ \mathbb{E}[Z] = \sqrt{\frac{2}{\pi}} $ leading to the claimed result. Similarly in order to compute variance, notice that $ \mathbb{E}[Y^{2}] = 0.5\mathbb{E}[(aZ)^{2}] + 0.5\mathbb{E}[Z^{2}] = 0.5\mathbb{E}[Z^{2}](1+a^{2})$. Then,
		\begin{equation}
		\begin{split}
		\var(Y) = \mathbb{E}[Y^{2}] - \mathbb{E}[Y]^{2} = 0.5\mathbb{E}[Z^{2}](1+a^{2}) - (1-a)^{2}\frac{1}{4}\mathbb{E}[Z]^{2}  \\
		= 0.5(1+a^{2})(\var(Z) + \mathbb{E}[Z]^{2})  - (1-a)^{2}\frac{1}{4}\mathbb{E}[Z]^{2} \\
		\end{split}
		\end{equation}
		Substituting $ \var(Z) =  1-\frac{2}{\pi}$ yields the claimed result.
	\end{proof}
\end{remark}

\end{appendix}